\def\year{2020}\relax
\documentclass[letterpaper]{article} % DO NOT CHANGE THIS
\usepackage{aaai20}  % DO NOT CHANGE THIS
\usepackage{times}  % DO NOT CHANGE THIS
\usepackage{helvet} % DO NOT CHANGE THIS
\usepackage{courier}  % DO NOT CHANGE THIS
\usepackage[hyphens]{url}  % DO NOT CHANGE THIS
\usepackage{graphicx} % DO NOT CHANGE THIS
\usepackage{graphicx}  % DO NOT CHANGE THIS
\usepackage{amsfonts,algorithmic,appendix}
\usepackage[linesnumbered,ruled]{algorithm2e}
\usepackage{amsmath, amssymb, amsthm}
\usepackage{booktabs,bm}
\usepackage{color}
\usepackage{epsf}
\usepackage{graphicx}
\usepackage{latexsym}
\usepackage{microtype}
\usepackage{subfigure}
\usepackage{textcomp,tikz}
\usepackage{url}
\usepackage{xcolor}
\usepackage{xspace}

\usetikzlibrary{arrows,shapes}
\usetikzlibrary{positioning}

\allowdisplaybreaks

\newtheorem{theorem}{Theorem}

\def\bfq{{{\mathbf q}}}

\def\bfx{{{\mathbf x}}}

\def\bfy{{{\mathbf y}}}

\def\bfz{{{\mathbf z}}}

\def\E{\ensuremath{{\mathbb E}}\xspace}

\def\R{\ensuremath{{\mathbb R}}\xspace}
\def\V{\ensuremath{{\mathbb V}}\xspace}

\def\argmin{\ensuremath{\mbox{argmin}}}

\def\BibTeX{{\rm B\kern-.05em{\sc i\kern-.025em b}\kern-.08em
    T\kern-.1667em\lower.7ex\hbox{E}\kern-.125emX}}

\title{Optimal Attack against Autoregressive Models by Manipulating the Environment}
\author{Yiding Chen, Xiaojin Zhu \\
Department of Computer Sciences, University of Wisconsin-Madison\\
\{yiding, jerryzhu\}@cs.wisc.edu
}
\begin{document}

\maketitle

\begin{abstract}
We describe an optimal adversarial attack formulation against autoregressive time series forecast using 
Linear Quadratic Regulator (LQR).  In this threat model, the environment evolves according to a dynamical system;  an autoregressive model observes the current environment state and predicts its future values; an attacker has the ability to modify the environment state in order to manipulate future autoregressive forecasts.  The attacker's goal is to force autoregressive forecasts into tracking a target trajectory while minimizing its attack expenditure.  In the white-box setting where the attacker knows the environment and forecast models, we present the optimal attack using LQR for linear models, and Model Predictive Control (MPC) for nonlinear models.  In the black-box setting, we combine system identification and MPC.  Experiments demonstrate the effectiveness of our attacks.
\end{abstract}

\section{Introduction}
Adversarial learning studies vulnerability in machine learning, see e.g.~\cite{vorobeychik2018adversarial,joseph2018adversarial,liu2017robust,DBLP:journals/corr/abs-1712-03141,lowd2005adversarial}.
Understanding optimal attacks that might be carried out by an adversary is important, as it prepares us to manage the damage and helps us develop defenses.
Time series forecast, specifically autoregressive model, is widely deployed in practice~\cite{hamilton1994time,box2015time,fan2008nonlinear}
but has not received the attention it deserves from adversarial learning researchers.
Adversarial attack in this context means an adversary can subtly perturb a dynamical system at the current time, hence influencing the forecasts about a future time.
Prior work~\cite{Alfeld2016Data,Alfeld2017Explicit} did point out vulnerabilities in autoregressive models under very specific attack assumptions.
However, it was not clear how to formulate general attacks against autoregressive models. 

There are extensive studies on \textbf{batch} adversarial attacks against machine learning algorithms. But there is much less work on \textbf{sequential} attacks. 
We say an attack is {batch} if the attacker performs one attack action at training or test time (the attacker is allowed to change multiple data points);
an attack is {sequential} if the attacker take actions over time.
There are {batch} attacks against support vector machine~\cite{biggio2012poisoning,biggio2014security}, deep neural networks~\cite{goodfellow2014explaining,nguyen2015deep}, differentially-private learners~\cite{ma2019data}, contextual bandits~\cite{Ma2018Data}, recurrent neural networks~\cite{papernot2016crafting}, online learning~\cite{wang2018data} and reinforcement learning~\cite{ma2019policy}. 
Some of these victims are sequential during deployment, but they can be trained from batch offline data; hence they can be prone to batch attacks.
In contrast, \cite{Jun2018Adversarial} and~\cite{zhang2019online} study sequential attacks against stochastic bandits and sequential prediction. 
Our work studies sequential attack against autoregressive model, which is closer to these two papers.
Meanwhile, control theory is receiving increasing attention from the adversarial learning community~\cite{recht2018tour,zhu2018optimal,lessard2018optimal}. Our work strengthens this connection.

This paper makes three main contributions: 
(1) 
We present an attack setting where the adversary must determine the attack sequentially. 
This generalizes the setting of~\cite{Alfeld2016Data,Alfeld2017Explicit}, where the adversary can decide the attack after observing all environmental state values used for forecast.
(2) We formulate the attacks as an optimal control problem.
(3) When the attacker knows the environmental dynamics and forecast model (white-box setting), we solve the optimal attacks with Linear Quadratic Regulator (LQR) for the linear case, or Model Predictive Control (MPC) and iterative LQR (iLQR) for the nonlinear case; when the attacker does not know the environmental or forecaster(black-box setting), we additionally perform system identification.  

\section{The Attack Setting}

\subsection{Autoregressive Review}
To fix notation, we briefly review time series forecasting using autoregressive models.  There are two separate entities:  

1. The \textbf{environment} is a fixed dynamical system with scalar-valued states $x_t \in \R$ at time $t$. 
%We consider scalar-valued states in this paper for simplicity; the extension to vector-valued states is straightforward.  
The environment has a 
(potentially non-linear) 
$q$-th order 
dynamics $f$ and is subject to zero-mean noise $w_t \in \R$ with $\V(w_t) = \sigma^2$.
Without manipulations from the adversary, the environmental state evolves as 
\begin{equation}
x_{t+1} = f(x_t, \ldots, x_{t-q+1},w_t)
\label{eq:fx}
\end{equation}
for $t=0,1,\ldots$.  We take the convention that $x_i=0$ if $i<0$.
We allow the dynamics $f$ to be either linear or nonlinear. 

%In white box setting, for linear dynamics we will solve the optimal adversarial attack with Riccati equation in LQR, while for nonlinear dynamics we will approximate the optimal attack using MPC / iLQR; in the black box setting, we propose system identification to model the environmental dynamics using linear function.

2. The \textbf{forecaster} makes predictions of future environmental states, and will be the victim of the adversary attack. In this paper we mainly focus on a fixed linear $AR(p)$ autoregressive forecaster, regardless of whether the environment dynamics $f$ is linear or not. Even though we allow the forecast model to be nonlinear in black-box setting, we use linear function to approximate the nonlinear autoregressive model.  We also allow the possibility $p \neq q$.  At time $t$, the forecaster observes $x_t$ and uses the $p$ most recent observations $x_t, \ldots, x_{t-p+1}$ to forecast the future values of the environmental state.
A forecast is made at time $t$ about a future time $t'>t$, we use the notation $y_{t' \mid t}$ to denote it.

Specifically, at time $t$ the forecaster uses a standard $AR(p)$ model to predict.
It initializes by setting $y_{t+1-i \mid t} = x_{t+1-i}$ for $i=1 \ldots p$.
It then predicts the state at time $t+1$ by
\begin{equation}
y_{t+1 \mid t} = \hat \alpha_0 + \sum_{i=1}^p \hat \alpha_i y_{t+1-i \mid t},
\label{eq:ARp}
\end{equation}
where $\hat \alpha_0, \hat \alpha_1, \ldots, \hat \alpha_p$ are coefficients of the $AR(p)$ model. 
We allow the $AR(p)$ model to be a nonlinear function in the black-box setting.
The $AR(p)$ model may differ from the true environment dynamics $f$ even when $f$ is linear: for example, the forecaster may have only obtained an approximate model from a previous learning phase.
Once the forecaster predicts $y_{t+1 \mid t}$, it can plug the predictive value in~\eqref{eq:ARp}, shift time by one, and predict $y_{t+2 \mid t}$, and so on.  Note all these predictions are made at time $t$.
In the next iteration when the true environment state evolves to $x_{t+1}$ and is observed by the forecaster, the forecaster will make predictions $y_{t+2 \mid t+1}, y_{t+3 \mid t+1}$, and so on.

\subsection{The Attacker}
We next introduce a third entity -- an \textbf{adversary} (a.k.a. attacker) -- who wishes to control the forecaster's predictions for nefarious purposes.
The threat model is characterized by three aspects of the adversary:

($i$) {Knowledge}: In the white-box setting, the attacker knows everything above; in the black-box setting, neither environmental dynamics nor forecaster model are known to the attacker.

($ii$) {Goal}: The adversary wants to force the forecaster's predictions $y_{t' \mid t}$ to be close to some given \emph{adversarial reference target} $y_{t' \mid t}^\dagger$ (the dagger is a mnemonic for attack), for selected pairs of $(t, t')$ of interest to the adversary.  Furthermore, the adversary wants to achieve this with ``small attacks''.  These will be made precise below. 
% Another possible setting is: the adversary wants to force the forecaster's prediction to be close to some given adversarial reference target and have small effect on the environment. 

($iii$) {Action}:
At time $t$ the adversary can add $u_t \in \R$ (the ``control input'') to the noise $w_t$.  Together $u_t$ and $w_t$ enter the environment dynamics via: 
\begin{equation}
x_{t+1} = f(x_t, \ldots, x_{t-q+1}, u_t+w_t).
\label{eq:fxu}
\end{equation}
We call this the \emph{state attack} because it changes the underlying environmental states, see Figure~\ref{fig:gm}.

\tikzstyle{state} = [draw, thin, minimum height=2em]
\tikzstyle{predictions} = [rounded corners, draw, thin, minimum height=2em]

\pgfdeclarelayer{background}
\pgfdeclarelayer{foreground}
\pgfsetlayers{background,main,foreground}

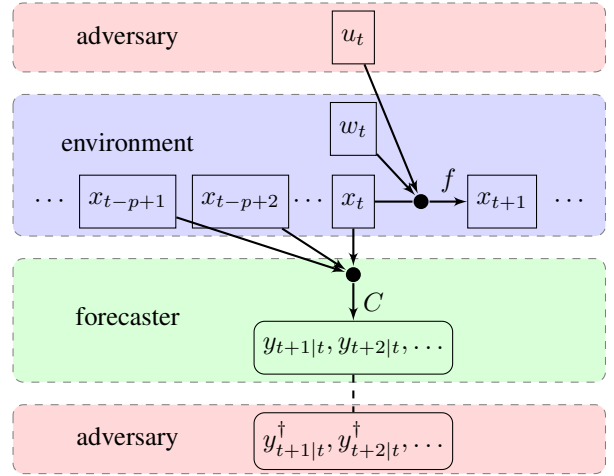
\begin{figure}[htp]
\centering
\begin{tikzpicture}[node distance=1.5cm, auto,>=latex', thick]

\node[state, draw = none] (0) {};
\node[state, right of = 0] (x1) {$x_{t-p+1}$} edge[draw = none] (0) ;
\node[state, right of = x1] (x2) {$x_{t-p+2}$};
\node[state, right of = x2] (x3) {$x_{t}$};
\node[circle,fill,inner sep=2pt, right = 0.5cm of x3] (d) {};
\node[state, right = 0.5cm of d] (x4) {$x_{t+1}$};
\node[state, draw = none, right of = x4] (1) {};

\node[state, draw = none, above = 0.25cm of x1] (2s) {};
\node[state, draw = none, above = 0.5cm of 2s] (1s) {adversary};
\node[circle,inner sep=2pt, draw = none, below = 0.5cm of x1] (4s) {};
\node[predictions, draw = none, below = 0.5cm of 4s] (5s) {};
\node[predictions, draw = none, below = 0.5cm of 5s] (6s) {adversary};

\node[state, draw = none, above = 0.075cm of x1] (env) {environment};
\node[predictions, draw = none, above = -0.3cm of 5s] (fore) {forecaster};

\node[state, above = 0.25cm of x3] (wt) {$w_t$};
\node[state, above = 0.5cm of wt] (ut) {$u_t$};

\node[circle,fill,inner sep=2pt, below = 0.5cm of x3] (d2) {};
\node[predictions, below = 0.5cm of d2] (y) {$y_{t+1 \mid t}, y_{t+2 \mid t}, \ldots$};
\node[predictions, below = 0.5cm of y] (yd) {$y_{t+1 \mid t}^\dagger, y_{t+2 \mid t}^\dagger, \ldots$};

\draw[->] (ut) edge (d);
\draw[->] (wt) edge (d);
\draw[->] (x1) edge (d2);
\draw[->] (x2) edge (d2);
\draw[->] (x3) edge (d2);
\draw[->] (d2) edge node[right] {$C$}(y);

\path (0) -- node[auto=false]{$\cdots$} (x1);
\path (x2) -- node[auto=false]{$\cdots$} (x3);
\path (x3) edge (d);
\draw[->] (d) edge node[above] {$f$}(x4);
\path (x4) -- node[auto=false]{$\cdots$} (1);

%\path (x1) -- node[auto = false] {environment} (2s);
%\path (4s) -- node[auto = false] {forecaster} (5s);

\draw (y) edge[dashed] (yd);

\begin{pgfonlayer}{background}
% Compute a few helper coordinates
\path (0.west |- 1s.north)+(0.1,0.1) node (a) {};
\path (1s.south -| 1.east)+(-0.25,-0.1) node (b) {};
\path[fill=red!15,rounded corners, draw=black!50, dashed] (a) rectangle (b);

\path (0.west |- 2s.north)+(0.1,0.1) node (a) {};
\path (1.south -| 1.east)+(-0.25,-0.1) node (b) {};
\path[fill=blue!15,rounded corners, draw=black!50, dashed] (a) rectangle (b);

\path (0.west |- 4s.north)+(0.1,0.1) node (a) {};
\path (5s.south -| 1.east)+(-0.25,-0.1) node (b) {};
\path[fill=green!15,rounded corners, draw=black!50, dashed] (a) rectangle (b);

\path (0.west |- 6s.north)+(0.1,0.1) node (a) {};
\path (6s.south -| 1.east)+(-0.25,-0.1) node (b) {};
\path[fill=red!15,rounded corners, draw=black!50, dashed] (a) rectangle (b);
\end{pgfonlayer}

\end{tikzpicture}
\caption{The state attack. The lowest layer depicts the attack target, and the adversary compares it against the forecaster's predictions.}
\label{fig:gm}
%\begin{minipage}{0.5\textwidth}
%\begin{center}
%\includegraphics[width=1\textwidth]{graphicalmodel2.png}
%\end{center}
%\caption{The observation attack, where the adversary manipulates what the forecaster sees but not the environment evolution.}
%\label{fig:gm2}
%\end{minipage}
\end{figure}

\section{White-Box Attack as Optimal Control \label{sec:formulate}}
We now present an optimal control formulation for the white-box attack.
Following control convention~\cite{lee1967foundations,kwakernaak1972linear}, we rewrite several quantities from the previous section in matrix form,
so that we can 
define the adversary's {state attack} problem by the tuple
$(F, C, \{\bfy_{t'\mid t}^\dagger\}, \{Q_{t'\mid t}\}, R, T)$.
These quantities are defined below.

We introduce a vector-valued environment state representation (denoted by boldface)
$
%\begin{equation}
\bfx_t := (1, x_{t}, \ldots, x_{t-p+1})^\top \in \R^{p+1}.
%\end{equation}
$
The first entry $1$ serves as an offset for the constant term $\hat \alpha_0$ in~\eqref{eq:ARp}.
We let $\bfx_0$ be the known initial state.  It is straightforward to generalize to an initial distribution over $\bfx_0$, which adds an expectation in~\eqref{eq:opt}.
We rewrite the environment dynamics under adversarial control~\eqref{eq:fxu} as
$
\bfx_{t+1} = F(\bfx_t, u_t, w_t) := (1, f(x_t, \ldots, x_{t-p+1}, u_t+w_t), x_t, \ldots, x_{t-p+2})^\top.
$
If $f$ is nonlinear, so is $F$.

In control language, the forecasts are essentially measurements of the current state $\bfx_t$.
In particular, we introduce a vector of $p$ predictions made at time $t$ about time $t'-p+1, \ldots, t'$ as
$
\bfy_{t' \mid t} :=
(1,
y_{t' \mid t},
y_{t'-1 \mid t},
\ldots
y_{t'-p+1 \mid t})^\top \in \R^{p+1}.
$
(For completeness, we let $y_{t' \mid t} = x_{t}$ when $t' \le t$.)  
The forecaster's $AR(p)$ forecast model specifies the (linear) measurements as follows.
We introduce the $(p+1) \times (p+1)$ measurement (i.e. forecast) matrix
%\footnote{
%The forecaster usually only has an estimate of the environment dynamics $f$.
%It is likely that the forecaster's $AR(p)$ model has a different order $p$ than the environment dynamics' order $q$.
%This can be handled in a straightforward manner, but the notation becomes cumbersome.
%Thus we will assume $p=q$ below, but explain how to handle $p \neq q$ in Appendix A.
%}
\footnote{
The forecaster usually only has an estimate of $f$.
It is likely that the forecaster's $AR(p)$ model has a different order $p$ than the environment dynamics' order $q$.
For simplicity, we will assume $p=q$ below, but explain how to handle $p \neq q$ in Appendix A.
}
,
\begin{equation}
\begin{small}
C = \begin{bmatrix}
1&0&0&\cdots&0&0\\
\hat \alpha_0 &\hat \alpha_1 & \hat \alpha_2&\cdots&\hat \alpha_{p-1}&\hat \alpha_p\\
0&1&0&\cdots&0&0\\
0&0&1&\cdots&0&0\\
\vdots&\vdots&\vdots&\ddots&\vdots&\vdots\\
0&0&0&\cdots&1&0\\
\end{bmatrix}.
\end{small}
\label{eq:hatA}
\end{equation}
Then the measurements / forecasts are:
\begin{equation}
\bfy_{t' \mid t} = C^{t'-t} \bfx_{t}, \quad \forall t' > t.
\label{eq:ARpmatrix}
\end{equation}
There are cases that $y_{t' \mid t}$ does not depend on $x_t$ but has been decided at time $\tilde t < t$. 
In such cases, we can simply redefine $y_{t' \mid t}$ to be $y_{t' \mid \tilde t}$ and rewrite the prediction matrix. 
For simplicity, we assume $y_{t' \mid t}$ always depends on $x_t$ in the rest of the paper.

We also vectorize adversarial reference target:
$
\bfy_{t' \mid t}^\dagger :=
(1,
y_{t' \mid t}^\dagger,
y_{t'-1 \mid t}^\dagger,
\ldots
y_{t'-p+1 \mid t}^\dagger)^\top \in \R^{p+1}.
$
We simply let $y_{t' \mid t}^\dagger = 0$ when $t' \le t$: this is non-essential.  In fact, for $(t,t')$ pairs that are uninteresting to the adversary, the target value $y_{t' \mid t}^\dagger$ can be undefined as they do not appear in the control cost later.

The cost of the adversary consists of two parts: (1) how closely the adversary can force the forecaster's predictions to match the adversarial reference targets; (2) how much control $u_t$ the adversary has to exert.
The $(p+1)\times(p+1)$ matrices $Q_{t'\mid t}$ define the first part, namely the cost of the attacker failing to achieve reference targets.
In its simplest form,  $Q_{t'\mid t}$ is a $(p+1)\times(p+1)$ matrix with all zero entries except for a scalar weight $\beta_{t' \mid t} \in [0, 1]$ at (2,2).
%\begin{equation}\label{eq:Qmatrix}
%Q_{t'\mid t}
%=
%\begin{bmatrix}
%0 & 0 & 0 & \cdots & 0\\
%0 & \beta_{t'\mid t} & 0 & \cdots & 0\\
%0 & 0 & 0 & \cdots & 0\\
%\vdots & \vdots & \vdots & \ddots & \vdots\\
%0 & 0 & 0 & \cdots & 0
%\end{bmatrix},
%\end{equation}
In this case, $Q_{t'\mid t}$ simply picks out the $(t' \mid t)$ element: 
\begin{equation}\label{eq:defQ}
( \bfy_{t' \mid t} - \bfy_{t' \mid t}^\dagger)^\top Q_{t'\mid t}( \bfy_{t' \mid t} - \bfy_{t' \mid t}^\dagger)
= \beta_{t' \mid t} ( y_{t' \mid t} - y_{t' \mid t}^\dagger)^2.
\end{equation}
For simplicity, we use $\| \bfy_{t' \mid t} - \bfy_{t' \mid t}^\dagger\|_{Q_{t'\mid t}}^2$ to denote~\eqref{eq:defQ}. 
Critically, by setting the weights $\beta_{t' \mid t}$ the attacker can express different patterns of attack.
For example:
% {\color{red} Somewhere in here we should place Scott's paper, this shows that our formulation generalizes that work.}
\begin{itemize}
%\item If $\beta_{n \mid m}=1$ for a specific pair $(m,n)$ and 0 for all other $t,t'$ pairs, the adversary is only interested in forcing the forecaster's prediction $y_{n \mid m}$ made at time $m$ about future time $n$ to be close to the adversarial reference target $y_{n \mid m}^\dagger$.
%The attacker does not care what happens to the other predictions.

\item If $\beta_{T \mid t}=1$ for all $t=0 \ldots T-1$ and 0 otherwise, the adversary cares about the forecasts made at all times about the final time horizon $T$. 
In this case, it is plausible that the adversarial target $y_{T \mid t}^\dagger := y_T^\dagger$ is a constant w.r.t. $t$. 

\item If $\beta_{t+1 \mid t}=1$ for all $t$ and 0 otherwise, the adversary cares about all the forecasts about ``tomorrow.''

\item if $\beta_{t' \mid t} = 1, \forall t,t'$, the adversary cares about all predictions made at all times.

\end{itemize}
Obviously, the adversary can express more complex temporal attack patterns.
The adversary can also choose $\beta_{t' \mid t}$ value in between $0$ and $1$ to indicate weaker importance of certain predictions.

The $R$ matrix defines the second part of the adversary cost, namely how much control expenditure $u_t$ the adversary has to exert.
In the simplest case, we let $R$ be a scalar $\lambda>0$:
\begin{equation}
\|u_t\|_R^2 := \lambda u_t^2.
\end{equation}

We use $T$ to denote the \emph{prediction time horizon}: $T$ is the last time index (expressed by $t'$) to be predicted by the forecaster.
%Given a time horizon $T$, 
We define the adversary's \emph{expected quadratic cost} $J(u_{0:(T-1)})$ for action sequence $u_0, \ldots, u_{T-1}$ by
%\footnote{
%A word is in order about the unusual summation range of $t$ (from 1 to $T-1$) in~\eqref{eq:quadraticcost}.
%We use $T$ to denote the \emph{prediction time horizon}: $T$ is the last time index (expressed by $t'$) to be predicted by the forecaster.
%Due to the fact that the forecaster always predicts the future, namely $t' > t$ in $y_{t'\mid t}$ (see the text below~\eqref{eq:ARp}), the last value for $t$ is therefore $T-1$.
%On the other end, although it is perfectly fine to talk about the forecasts $y_{t' \mid t=0}$ made at the very beginning about a future time $t'$, such zero-time predictions are not controllable by the adversary: the adversary cannot change $\bfx_0$ and thus cannot change the forecaster's predictions made on time $t=0$.
%In other words, $\forall t', y_{t' \mid t=0}$ are constants with respect to the control policies $\phi_0, \ldots, \phi_{T-1}$.
%As such, we remove them from the adversary's objective function~\eqref{eq:quadraticcost}.
%This explains why in the first cost term $t$ starts from 1.
%We emphasize that the above discussion only concerns the cost function; the adversary seeks control policies $\phi_t$ for $t=0 \ldots T-1$.
%}
\begin{equation*}
%\begin{small}
\E_{w_{0:(T-1)}} \! \! \left[ \left. \sum_{t=1}^{T-1}\sum_{t' = t+1}^{T} \! \|\bfy_{t' \mid t} - \bfy_{t' \mid t}^\dagger\|^2_{Q_{t' \mid t}} \! + \! \sum_{t=0}^{T-1} \|u_t\|_R^2 \right| \bfx_0 \right].
\label{eq:quadraticcost}
%\end{small}
\end{equation*}

Since the environment dynamics can be stochastic, the adversary must seek \emph{attack policies} $\phi_t: \R^{p+1} \mapsto \R$ to map the observed state $\bfx_t$ to an attack action:
\begin{equation}
u_t = \phi_t(\bfx_t) \quad \forall t.
\end{equation}

Given an adversarial {state attack problem} $(F, C, \{\bfy_{t'\mid t}^\dagger\}, \{Q_{t'\mid t}\}, R, T)$,
we formulate the optimal state attack as the following optimal control problem:
\begin{align}
&\min_{\phi_0,\cdots,\phi_{T-1}}  J(u_{0:(T-1)}) \label{eq:opt} \\
\mbox{s.t.} & \bfx_0 \mbox{ given} \\
	& u_t = \phi_t(\bfx_t), \quad t = 0,1,\cdots,T-1 \\
	& \bfx_{t+1} = F(\bfx_t, u_t, w_t),\quad t = 0,1,\cdots,T-1 \\
	& \bfy_{t' \mid t} = C^{t'-t} \bfx_{t}, \quad \forall t' > t.
\end{align}
We next propose solutions to this control problem for linear $F$ and nonlinear $F$, respectively. For illustrative purpose, we focus on solving the problem when the attack target is to change predictions for ``tomorrows''. This implies $\beta_{t' \mid t} = 0$ when $t' \ge t+2$. Under this assumption, $J(u_{0:(T-1)})$ has the following form:
\begin{equation*}
\E_{w_{0:(T-1)}} \left[ \left. \sum_{t=1}^{T-1} \|C \bfx_t - \bfy_{t+1 \mid t}^\dagger\|^2_{Q_{t+1 \mid t}} + \sum_{t=0}^{T-1} \|u_t\|_R^2 \right| \bfx_0 \right].
\end{equation*}
More attack targets are studied in the experiment section. 

\subsection{Solving Attacks Under Linear $F$}
When the environment dynamics $f$ is linear, the scalar environment state evolves as 
$
x_{t+1} = \alpha_0 + \sum_{i=1}^p \alpha_i x_{t+1-i} + u_t + w_t, 
$%\label{eq:ARq}
where the coefficients $\alpha_0, \ldots, \alpha_p$ in general can be different from the forecaster's $AR(p)$ model~\eqref{eq:ARp}.
We introduce the corresponding vector operation 
\begin{equation}
\bfx_{t+1} = F(\bfx_t, u_t, w_t) := A \bfx_t + B (u_t +  w_t),
\end{equation}
where $A$ has the same structure as $C$ in~\eqref{eq:hatA} except each $\hat\alpha$ is replaced by $\alpha$,
% \begin{equation}
% \begin{small}
% A = \begin{bmatrix}
% 1&0&0&\cdots&0&0\\
% \alpha_0 &\alpha_1 & \alpha_2&\cdots&\alpha_{q-1}&\alpha_p\\
% 0&1&0&\cdots&0&0\\
% 0&0&1&\cdots&0&0\\
% \vdots&\vdots&\vdots&\ddots&\vdots&\vdots\\
% 0&0&0&\cdots&1&0\\
% \end{bmatrix}
% \end{small}
% \end{equation}
and $B=(0,1,0,\ldots,0)^\top$.
The adversary's attack problem~\eqref{eq:opt} reduces to stochastic Linear Quadratic Regulator (LQR) with tracking, which is a fundamental problem in control theory~\cite{kwakernaak1972linear}. 
% {\color{red} cite XXX}
It is well known that such problems have a closed-form solution, though the specific solution for stochastic tracking is often omitted from the literature. 
In addition, the presence of a forecaster in our case alters the form of the solution.
Therefore, for completeness we provide the solution in Algorithm~\ref{alg:LQR}.
\begin{algorithm}
\SetKwInOut{Input}{Input}
\SetKwInOut{Output}{Output}
\caption{$LQR(F, C, \{\bfy_{t'\mid t}^\dagger\}, \{Q_{t'\mid t}\}, R, T)$\label{alg:LQR}}
\Input{$(F, C, \{\bfy_{t'\mid t}^\dagger\}, \{Q_{t'\mid t}\}, R, T)$}
$P_T = 0$\;
$\bfq_T = 0$\;
\For{$t = T-1, T-2, \cdots,1$}{
$P_{t} = C A^\top Q_{t+1 \mid t} C + A^\top (I + \frac{1}{\lambda}P_{t+1} BB^\top)^{-1}P_{t+1} A$\;
$\bfq_{t} = -2 C^\top Q_{t+1 \mid t} \bfy_{t+1 \mid t}^\dagger + A^\top \bfq_{t+1} \nonumber -\frac{1}{\lambda + B^\top P_{t+1} B} A^\top P_{t+1}^\top BB^\top \bfq_{t+1}$\;
}
\For{$t=0,1,\ldots,T-1$}{
$\phi_t (\bfz) =  -\frac{B^\top \bfq_{t+1} + 2 B^\top P_{t+1} A \bfz}{2(\lambda + B^\top P_{t+1} B)}$
}
\Output{$\phi_{0}(\cdot),\ldots,\phi_{T-1}(\cdot)$}
\end{algorithm}

The derivation is in Appendix.
Once the adversarial control policies are computed, the optimal attack sequence is given by:
$
u_t = \phi_t(\bfx_t), \quad t = 0,1,\cdots,T-1.
$
The astute reader will notice that, 
$u_{T-1} = \phi_{T-1}(\bfx_{T-1})=0$.
This is to be expected: $u_{T-1}$ affects $\bfx_T$, but $\bfx_T$ would only affect forecasts after the prediction time horizon $T$, which the adversary does not care. 
To minimize the control expenditure, the adversary's rational behavior is to set $u_{T-1}=0$.

\subsection{Solving Attacks Under Non-Linear $F$}
When $f$ is nonlinear the optimal control problem~\eqref{eq:opt} in general does not have a closed-form solution. 
Instead, we introduce an algorithm that combines Model Predictive Control (MPC)~\cite{garcia1989model,kouvaritakis2015stochastic} as the outer loop  and Iterative Linear Quadratic Regulator (ILQR)~\cite{li2004iterative} as the inner loop to find an approximately optimal attack. %{\color{red} cite ILQR, Stochastic Model Predictive Control (Basil Kouvaritakis), Input-to-state stable MPC for constrained discrete-time nonlinear systems with bounded additive uncertainties}
While these techniques are standard in the control community, to our knowledge our algorithm is a novel application of the techniques to adversarial learning.

The outer loop performs MPC, a common heuristic in nonlinear control.
At each time $\tau=0,1,\ldots$, MPC performs planning by starting at $\bfx_\tau$, looking ahead $l$ steps and finding a good control sequence $\{\phi_t\}^{\tau+l-1}_{t = \tau}$.
However, MPC then carries out only the first control action $u^*_\tau$.
This action, together with the actual noise instantiation $w_\tau$, drives the environment state to $\bfx_{\tau+1} = F(\bfx_\tau, u^*_\tau, w_\tau)$.
Then, MPC performs the $l$-step planning again but starting at $\bfx_{\tau+1}$, and again carries out the first control action $u^*_{\tau+1}$.
This process repeats.
Formally, MPC iterates two steps: at time $\tau$

1. Solve 
\begin{align}
\min_{\phi_{\tau:L(\tau)}} & \E\sum_{t=\tau+1}^{L(\tau)+1} \|C \bfx_{t} - \bfy_{t+1 \mid t}^\dagger\|^2_{Q_{t+1 \mid t}} + \sum_{t=\tau}^{L(\tau)}\|\phi_t(x_t)\|_R^2  \nonumber\\
\text{s.t.} & \bfx_\tau \text{ given} \nonumber\\
& \bfx_{t+1} = F(\bfx_t, \phi_t(u_t), w_t), t = \tau,\cdots,L(\tau), \label{eq:mpc}
\end{align}

%where 
%\begin{equation}\label{eq:L}
%$L(\tau) = \min(\tau+l-1,T-2)$.
%\end{equation}
The expectation is over $w_\tau,\cdots,w_{L(\tau)}$. 
Denote the solution by $\{\phi_t\}^{L(\tau)}_{t = \tau}$.

2. Apply $u^*_\tau = \phi_\tau(x_\tau)$ to the $F$ system. 

$L(\tau)=\min(\tau+l-1,T-2)$, which indicates that the size of the optimization in step 1 will decrease as $\tau$ approaches $T$.
The repeated re-planning allows MPC to adjust to new inputs, and provides some leeway if $\{\phi_t\}^{L(\tau)}_{t = \tau}$ cannot be exactly solved, which is the case for our nonlinear $F$.

We now turn to the inner loop to approximately solve~\eqref{eq:mpc}.
There are two issues that make the problem hard: the expectation over noises $\E_{w_\tau,\cdots,w_{L(\tau)}}$, and the nonlinear $F$.
To address the first issue, we adopt an approximation technique known as ``nominal cost'' in~\cite{kouvaritakis2015stochastic}. %{\color{red} (cite ILQG, Stochastic Model Predictive Control (Basil Kouvaritakis))}. 
For planning we simply replace the random variables $w$ with their mean, which is zero in our case.  This heuristic removes the expectation, and we are left with the following deterministic system as an approximation to~\eqref{eq:mpc}:
\begin{align}%\label{eq:nmpc}
\min_{u_\tau,\cdots,u_{L(\tau)}} & \sum_{t=\tau+1}^{L(\tau)+1}\|C \bfx_{t} - \bfy_{t+1 \mid t}^\dagger\|^2_{Q_{t+1 \mid t}} + \sum_{t=\tau}^{L(\tau)}\|u_t\|_R^2 \nonumber\\
\text{s.t.} & \bfx_\tau \text{ given} \nonumber\\
& \bfx_{t+1} = F(\bfx_t, u_t, 0),\quad t = \tau,\cdots,L(\tau). \label{eq:nmpc}%\label{eq:deter}
\end{align}

To address the second issue, we adopt ILQR~\cite{li2004iterative} in order to solve~\eqref{eq:nmpc}. The idea of ILQR is to linearize the system around a trajectory, and compute an improvement to the control sequence using LQR iteratively. We show the details in Appendix. We summarize the MPC+ILQR attack in Algorithm~\ref{alg:MPC} and~\ref{alg:ILQR}.

\begin{algorithm}
\SetKwInOut{Input}{Input}
\SetKwInOut{Output}{Output}
\caption{MPC\label{alg:MPC}}
\Input{$F, C, \{\bfy_{t'\mid t}^\dagger\}, \{Q_{t'\mid t}\}, R, T, l, maxiter, tol$}
\For{$t=0,1,\ldots,T-2$}{
\Input{$x_t$}
$u_{t:\min(t+l-1,T-2)} \leftarrow ILQR(x_t, F, C, \{\bfy_{t'\mid t}^\dagger\}, \{Q_{t'\mid t}\}, R, \min(t+l+1,T), maxiter, tol)$\;
\Output{$u_t$}
}
\end{algorithm}

\begin{algorithm}[ht]
\SetKwInOut{Input}{Input}
\SetKwInOut{Output}{Output}
\caption{ILQR\label{alg:ILQR}}
\Input{$x_0, F, C, \{\bfy_{t'\mid t}^\dagger\}, \{Q_{t'\mid t}\}, R, T, maxiter, tol$}
Initialize $u_{0:T-2}$\;
\For{$i=0,1,\ldots,maxiter$}{
\For{$t = 0:T-2$}{
$\bfx_{t+1} = F(\bfx_{t},u_{t},0)$\;
$D_uF_t = D_uF(\bfx_t,u_t,0)$\;
$D_xF_t = D_xF(\bfx_t,u_t,0)$
}
$P_{T-1} = C^\top Q_{T \mid T-1} C$\;
$\bfq_{T-1} = 2C^\top Q_{T \mid T-1} (C \bfx_{T-1}- \bfy_{T \mid T-1}^\dagger)$\;
\For{$s = T-2,\ldots,1$}{
$P_s = C^\top Q_{s+1 \mid s}C 
+ D_\bfx F_s^\top(I + \frac{1}{\lambda}P_{s+1} D_u F_s D_u F_s^\top)^{-1}P_{s+1}D_\bfx F_s^\top $\;
$\bfq_s = 2C^\top Q_{s+1 \mid s}(C \bfx_s- \bfy_{s+1 \mid s}^\dagger) + D_\bfx F_s^\top \bfq_{s+1} 
-\frac{ (D_u F_s^\top P_{s+1} D_\bfx F_s)^\top (D_u F_s^\top \bfq_{s+1} + 2 \lambda  u_s)  }{\lambda + D_u F_s^\top P_{s+1} D_u F_s}$
}
\For{$s=0,T-2$}{
$\delta u_s = -\frac{2 D_u F_s^\top P_{s+1} D_\bfx F_s\delta\bfx_s + D_u F_s^\top \bfq_{s+1} + 2 \lambda  u_s}{2(\lambda + D_u F_s^\top P_{s+1} D_u F_s)}$\;
$\delta\bfx_{s+1} = D_\bfx F_s \delta\bfx_s + D_u F_s \delta u_s$
}
\If{$\|\delta u_{0:T-2}\|^2/(T-1) < tol$}{
	Break\;
}
$u_{0:T-2} \leftarrow u_{0:T-2} + \delta u_{0:T-2}$\;
}
\Output{$u_{0:T-2}$}
\end{algorithm}

%When the optimal $\delta u^*$s is obtained, an "better" control sequence is found by: $\tilde u_t \leftarrow \delta u_t^* + \tilde u_t, t = \tau, \tau+1, \cdots, L(\tau)$.

\subsection{A Greedy Control Policy as the Baseline State Attack Strategy}
\label{sec:greedy}
The optimal state attack objective~\eqref{eq:opt} can be rewritten as a running sum of instantaneous costs. At time $t=0,1,\ldots$ the instantaneous cost involves the adversary's control expenditure $u_t^2$, the attack's immediate effect on the environment state $\bfx_{t+1}$ (see Figure~\ref{fig:gm}), and consequently on all the forecaster's predictions made at time $t+1$ about time $t+2, \ldots, T$.
Specifically, the expected instantaneous cost $g_t(\bfx_t, u_t)$ at time $t$ is defined as:
\begin{equation}
%g_t(\bfx_t, u_t) := 
\E_{w_t}\|C F(\bfx_t, u_t, w_t)- {\bfy}_{t+2 \mid t+1}^\dagger\|^2_{Q_{t+2 \mid t+1}} + \|u_t\|_R^2.
\end{equation}
This allows us to define a \emph{greedy control policy} $\phi^G$, which is easy to compute and will serve as a baseline for state attacks.
In particular, the greedy control policy at time $t$ minimizes the instantaneous cost:
$$
\phi^G_t(\bfx_t) \in \argmin_{u} g_t(\bfx_t, u).
$$
When $F$ is linear, $\phi^G_t(\cdot)$ can be obtained in closed-form.
We show the solution in Appendix D.

When $f$ is nonlinear, we let noise $w_t = 0$ and solve the following nonlinear problem using numerical solvers:
\begin{equation}\label{eq:nlgdob}
\min_{u_t}\|C F(\bfx_t, u_t, 0)- {\bfy}_{t+2 \mid t+1}^\dagger\|^2_{Q_{t+2 \mid t+1}} + \|u_t\|_R^2.
\end{equation}
%To fit this objective function into the solver, we need to first rewrite the objective function in~\eqref{eq:nlgdob}. Firstly, denote $\bar{\beta}_{t' \mid t} = (0, \sqrt{\beta_{t' \mid t}}, 0, \cdots, 0)^\top$. Then $Q_{t' \mid t} = \bar{\beta}_{t' \mid t} \bar{\beta}_{t' \mid t}^\top$. The objective function can be written as:
%\begin{equation}
%\sum_{t' = t+2}^T (\bar{\beta}_{t' \mid t+1}^\top (C^{t'-(t+1)}F(\bfx_t, u_t, 0)-\bar{\bfx}_{t' \mid t+1}))^2 + \lambda u_t^2.
%\end{equation}

\section{Black-Box Attack via System Identification}
We now consider black-box attack setting where the environment dynamics $f$ and forecaster's model $C$ are no longer known to the attacker. Both the environment forecast models are allowed to be nonlinear. 
The attacker will perform system identification~\cite{dean2017sample},  and solve LQR as an inner loop and MPC as an outer loop.

The attacks picks an estimation model order $p$ for both the environment and forecaster.
It also picks a buffer length $b$.
In the first $b+p-1$ iterations, the attacker does no attack but collects observations on the free-evolving environment and the forecasts. 
Then, in each subsequent iteration the attacker estimates a linear environmental model $\hat{\bm{a}}$ and linear forecast model $\hat{\bm{c}}$ using a rolling buffer of previous $b+p-1$ iterations. 
The buffer produces $b$ data points for the attacker to use MLE to solve $p+1$ unknowns in environment model.% and some data points depending on .
The attacker then uses MPC and LQR to design an attack action. 

%When a large $p$ is chosen, the attacker might get a better estimation but will suffer intense computation. 
The attacker use linear models to estimate both  the environmental dynamics and forecast model. At time $t$, he environmental dynamics is estimated over the environmental state value $x_{t-b-p+1}, \ldots, x_{t}$ and action sequences $u_{t-b}, \ldots, u_{t-1}$:
\begin{equation}
\min_{a_{0:p}} \sum_{t = t-b}^{t-1} (a_0 + \sum_{i=1}^p a_i x_{t+1-i} + u_t - x_{t+1})^2,
\label{eq:estA}
\end{equation}
we use $\hat{\bm{a}}(t, b)$, a $p+1$-dimensional column vector to denote the minimum point of~\eqref{eq:estA}.
Due to the nature of attacking autoregressive models, $B = (0,1,0,\ldots,0)^\top \in \R^{(p+1) \times 1}$ is always known to the attacker.

% If the forecaster has a general prediction pattern, the attacker might have to solve a system of polynomials of forecast model coefficient, which is hard in general, (see Appendix E) we leave this case as future work. In this paper, for simplicity, we assume that the attacker has enough $1$-ahead observations $y_{t+1 \mid t}$ to estimate the forecaster's model.

We use $\hat c_0, \hat c_1, \ldots, \hat c_p$ to denote the estimation of forecast model and let 
\begin{equation}
%\begin{small}
C(\hat c_{0:p}) = \begin{bmatrix}
1&0&0&\cdots&0&0\\
\hat c_0 &\hat c_1 & \hat c_2&\cdots&\hat c_{p-1}&\hat c_p\\
0&1&0&\cdots&0&0\\
0&0&1&\cdots&0&0\\
\vdots&\vdots&\vdots&\ddots&\vdots&\vdots\\
0&0&0&\cdots&1&0\\
\end{bmatrix}.
%\end{small}
\label{eq:hatC}
\end{equation}
to denote the corresponding forecast matrix.
Let $\Lambda(t_1, t_2)$ denote the set of prediction indices which are visible to the attacker: $\Lambda(t_1, t_2) = \{(t,t') \mid t_1 \le t \le t_2, y_{t' \mid t} \mbox{ is visible to the attacker}\}$.
At time $t$, the forecast model is estimated over the visible forecasts: $y_{t' \mid t}, (t,t') \in \Lambda(t-b, t)$:
\begin{equation}
\min_{c_{0:p}} \sum_{(t, t') \in \Lambda(t-b, t)}(B^\top C^{t' - t} \bfx_{t} - y_{t' \mid t})^2,
\label{eq:estC}
\end{equation}
we use $\hat{\bm{c}}(t, b)$, a $p+1$-dimensional column vector to denote the minimum point of~\eqref{eq:estC}.
If the attacker only observes sufficient predictions for ``tomorrows'', i.e. $\Lambda(t_1, t_2) = \{(t,t') \mid t_1 \le t \le t_2, t' = t+1\}$, then $\hat c_0, \hat c_1, \ldots, \hat c_p$ is the OLS solution to~\eqref{eq:estC}. 
However, for more complex prediction pattern,~\eqref{eq:estC} might involve polynomials of $c_0, c_1, \ldots, c_p$.

We can summarize the proposed black-box attack method in Algorithm~\ref{alg:sida}.
%\begin{enumerate}
%\item For $t = 0,1, \ldots, M$: apply $u_t = 0$ and observe $x_t$s and $y_{t' \mid t}$s;
%\item For $t = M+1, \ldots, T-1$: 
%\begin{itemize}
%\item update estimations for environmental  dynamics $\hat{\bm{a}}(t - L_1, t)$ and forecast model $\hat{\bm{c}}(t - L_1, t)$ by~\eqref{eq:estA} and ~\eqref{eq:estC};
%\item solve the LQR problem defined by $(x_t, \{\hat{\bm{a}}(t-L_1,t ), B\}, C(\hat{\bm{c}}(t - L_1, t)), \{\bfy_{t'\mid t}^\dagger\}, \{Q_{t'\mid t}\}, R, [t, \min(t+L_2, T)])$ ($x_t$ is the initial state of this LQR problem) and obtain $u_t, u_{t+1}, \ldots, u_{\min(t+L_2, T-1)-1}$;
%\item apply $u_t$ to the environment.
%\end{itemize}
%\end{enumerate}

\begin{algorithm}
\SetKwInOut{Input}{Input}
\SetKwInOut{Output}{Output}

\caption{System identification attack~\label{alg:sida}}
\Input{model order $p$, buffer size $b$, time step of MPC $l$}
\For{$t = 0,1, \ldots, b+p-2$}{
\Input{$x_t$, $y_{t' \mid t}$}
\Output{$u_t = 0$}
}
\For{$t = b+p-1, \ldots, T-1$}{
update $\hat{\bm{a}}(t, b)$ by~\eqref{eq:estA}\;
update $\hat{\bm{c}}(t, b)$ by~\eqref{eq:estC}\;
$\phi_{0:\min(t+l, T)-1} \leftarrow LQR(\{\hat{\bm{a}}(t,b ), B\}, C(\hat{\bm{c}}(t, b)), \{\bfy_{t'\mid t}^\dagger\}, \{Q_{t'\mid t}\}, R, \allowbreak \min(t+l+1, T)-t)$\;
\Output{$u_t \leftarrow \phi_0(x_t)$}
\Input{$x_{t+1}$, $y_{t' \mid t}$}
}
\end{algorithm}

%This attack methods includes the following parameters: $p$ is the model order; $M$ is a time step, before which the attacker simply observes environment state and predictions but applies no attack to the environment; $L_1$ characterizes a set of observations that will be used by the attacker to update the estimations for environmental dynamics and forecast model; $L_2$ is the length of MPC time step. 

\section{Experiments}
We now demonstrate the effectiveness of control-based white-box attacks on time series forecast problems.  
We compare the optimal attacks computed by LQR (for linear $f$), MPC+iLQR (for nonlinear $f$), black-box attack, greedy attacks, and the no-attack baseline.
While the attack actions were optimized under an expectation over random noise $w$ (c.f.~\eqref{eq:opt}), in the experiments we report the \emph{actual realized cost}
based on the noise instantiation that the algorithm experienced:
\begin{equation}
\sum_{t=1}^{T-1}\sum_{t' = t+1}^{T}\|C^{t'-t} \bfx_{t} - \bfy_{t' \mid t}^\dagger\|^2_{Q_{t' \mid t}} + \sum_{t=0}^{T-1}\|u_t\|_R^2
\end{equation}
where the noise sequence $\{w_t\}_{t=0}^{T-2}$ is incorporated implicitly in $\bfx_t$, together with the actual attack sequence  $\{u_t\}_{t=0}^{T-1}$. 
To make the balance between attack effect (the quadratic terms involving $Q$) and control expenditure (the term involving $R$) more interpretable, we let
$
R := \lambda = \tilde\lambda {\sum_{t = 1}^{T-1}\sum_{t'  = t+1}^T\beta_{t'\mid t}}/{T}
$

\subsection{The Effect of $Q_{t'\mid t}$ on Attack \label{sec:toy1}}

\begin{figure}[th]
\centering
\includegraphics[width=0.47\textwidth]{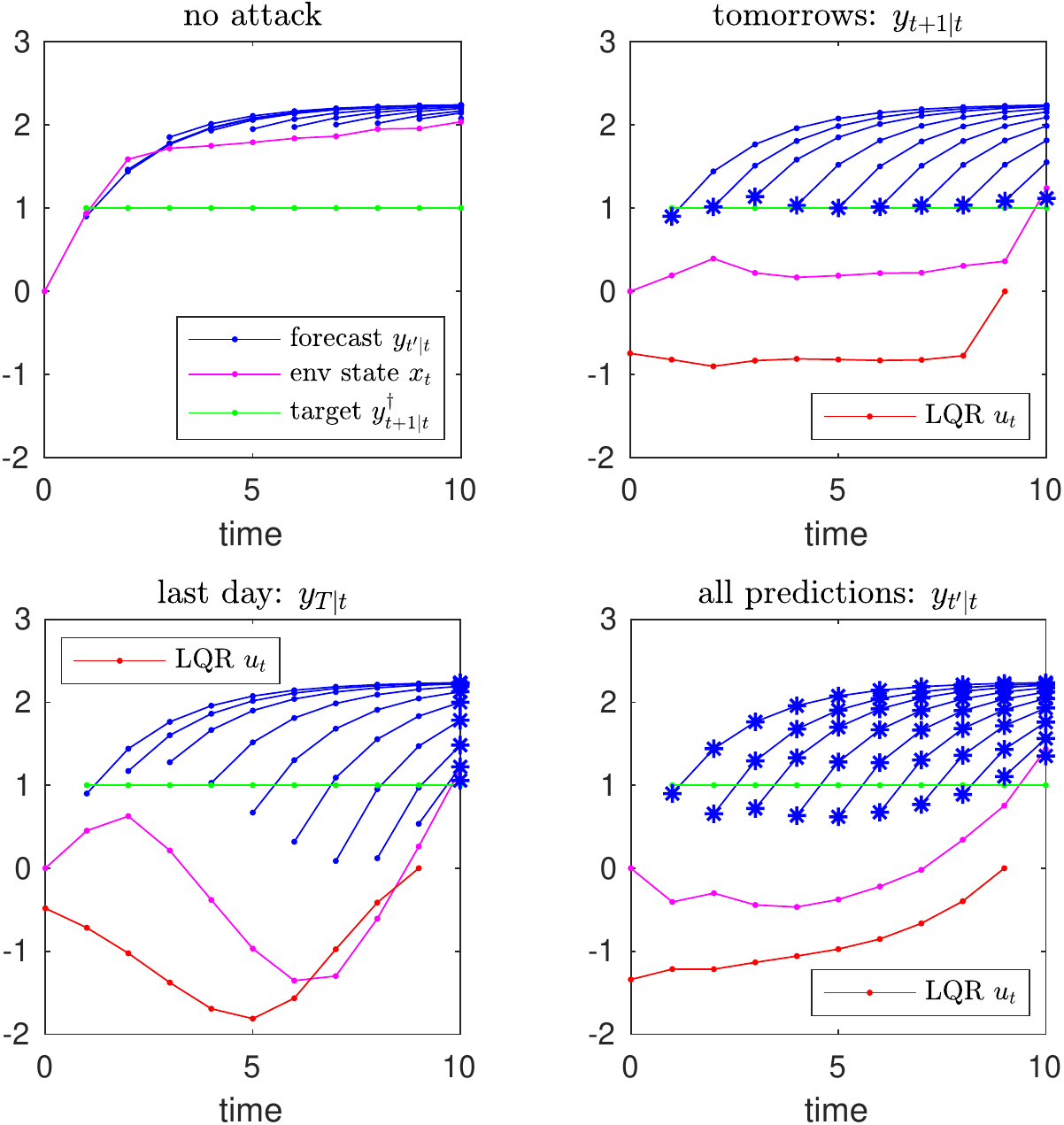}
\caption{
LQR solution on three attack patterns $Q$. $y$-axis shows the value of $x$ and $u$.  Each blue forecast curve beginning at time $t$ shows the sequence $\{y_{t'|t-1}\}_{t' = t}^{T}$. The pattern of attack defined by the corresponding $Q$ is highlighted with $*$ on the forecast curves.\label{fig:toy1}}
\end{figure}

In our first synthetic example we demonstrate \textbf{the adversary's ability to target different parts of the forecasts via $Q$},
the quadratic coefficient in cost function. 
Figure~\ref{fig:toy1} illustrates three choices of attack targets $Q$: attack ``tomorrow'', ``last day'' and ``all predictions''.

%1. ``Tomorrow'': The attacker only cares about the forecasts $y_{t+1 \mid t}$ made on $t=1, \ldots T-1$ about the immediate next time step.
%That is, at $t=1$ the attacker wants to force the forecast of time step 2, namely $y_{2 \mid 1}$, to be close to the adversarial reference target $ y_{2\mid 1}^\dagger$; at $t=2$ force the forecast of time step 3 $y_{3 \mid 2}$ to be close to $y_{3 \mid 2}^\dagger$, and so on.
%This is done by setting $\beta_{t+1 \mid t}=1$ in $Q_{t+1 \mid t}$ for $t=1, \ldots T-1$, and let all other $Q_{t'\mid t}$ to be zero.

%2. ``Last day'': the attacker only cares about predictions $y_{T \mid t}$ made on $t=1, \ldots T-1$. 

%3. ``All predictions'': 
%the attacker wants to match any prediction $y_{t' \mid t}$ to its reference target $ y_{t' \mid t}^\dagger$ for $t=1, \ldots T-1$ and $t'=t+1 \ldots T$.

For simplicity, we let all adversarial reference targets $y_{t'\mid t}^\dagger$ be the constant 1.
We let the environment evolve according to an $AR(1)$ model: 
$
x_{t+1} = 1+0.5 x_{t} + w_t.
$
%Equivalently, $A= 
%\begin{pmatrix}
%1&0\\
%1&0.5
%\end{pmatrix}$
%, $B = (0,1)^\top$ in matrix notation.
We let the noise $w_t \sim N(0,0.1^2)$ and the initial state $x_0 = 0$.
We simulate the case where the forecaster has only an approximation of the environment dynamics, and let the forecaster's model be 
$
x_{t+1} = 0.9 + 0.6 x_{t} 
$
which is close to, but different from, the environment dynamics.
%The corresponding forecast matrix is $C =
%\begin{pmatrix}
%1&0\\
%0.9&0.6
%\end{pmatrix}$.
For illustrative purpose, we set the prediction time horizon $T = 10$. 
Recall that the attacker can change the environment state by adding perturbation $u_t$: $x_{t+1} = 1+0.5 x_{t} + u_t + w_t$. We set $\tilde\lambda = 0.1$.

We run LQR and compute the optimal attack sequences $u$ under each $Q$ scenario.
They are visualized in Figure~\ref{fig:toy1}. 
Each attack is effective: the blue *'s are closer to the green target line on average, compared to where they would be in the upper-left no-attack panel. Different target selection $Q$ will affect the optimal attack sequence.

\subsection{Comparing LQR vs. Greedy Attack Policies}
We now show \textbf{the LQR attack policy is better than the greedy attack policy}.
We let the environment evolves by an $AR(3)$ model: 
$
x_{t+1} = 
f(x_t,x_{t-1}, x_{t-2}, w_t) = 0.4x_{t} -0.3x_{t-1} - 0.7 x_{t-2} + w_t
$
, $w_t\sim N(0,0.1^2)$. 
%Equivalently, $A= 
%\begin{pmatrix}
%1&0&0&0\\
%0&0.4&-0.3&-0.7\\
%0&1&0&0\\
%0&0&1&0
%\end{pmatrix}$, $B = (0,1,0)^\top$ in matrix notation. 
The initial values are $x_0=10$, $x_{-1} = x_{-2} = 0$, prediction horizon $T = 15$. 
This environment dynamic is oscillating around $0$.

We let the forecaster's model be: % a close approximation to the environment:
$
x_{t+1} = 0.41x_{t} -0.29x_{t-1} - 0.68 x_{t-2}. 
$
%Equivalently, $C= 
%\begin{pmatrix}
%1&0&0&0\\
%0&0.41&-0.29&-0.68\\
%0&1&0&0\\
%0&0&1&0
%\end{pmatrix}$ in matrix notation. 
%The attacker can change the state value by adding perturbation: $x_{t+1} = 0.4x_{t} -0.3x_{t-1} - 0.7 x_{t-1} + u_t + w_t$. The attacker wants to change predictions about "tomorrows" ($y_{t+1 \mid t}, t = 0,1,\cdots,T-1$). 
$Q$ is "tomorrows".
The attacker wants the forecaster to predict a sequence oscillating with smaller amplitude. $ y_{t+1 \mid t}^\dagger$ is set as following: 
we simulate $x_{t+1} = f(x_t,x_{t-1},x_{t-2},0)$, then, let the attack reference target be ${y}_{t \mid t-1}^\dagger = 0.5x_{t}, t = 2, \cdots, T$.  
%Again we let $\lambda = \tilde\lambda \frac{\sum_{t = 1}^{T-1}\sum_{t'  = t+1}^T\beta_{t'\mid t}}{T}$
We set $\tilde\lambda = 0.1$. 

\begin{figure}[htbp]
\centering
\includegraphics[width=0.47\textwidth]{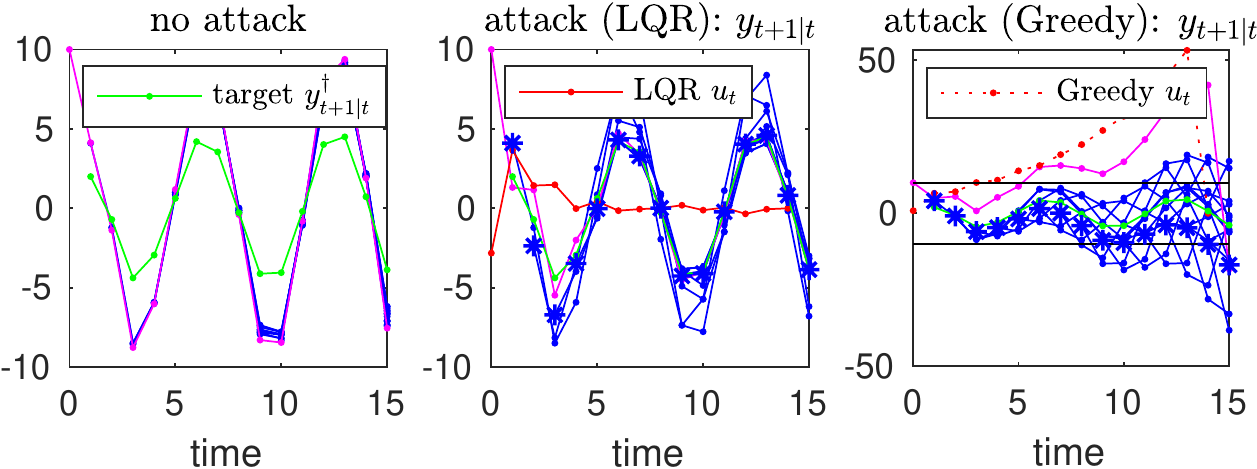}
\caption{LQR vs. Greedy attacks. The black horizontal lines in the right plot mark the vertical axis range of the middle plot.  \label{fig:toy2}}
\end{figure}

We run LQR and Greedy, respectively, to solve this attacking problem. We generate $50$ trials with different noise sequences,  see Figure~\ref{fig:toy2}. 
Interestingly, LQR drives the predictions close to the attack target, while Greedy \emph{diverges}.
The mean actual realized cost of no attack, LQR attack and Greedy attack are $133.9, 11.5, 1492$, respectively. The standard errors are $1.20,0.02,12.99$. We perform a paired $t$-test on LQR vs. Greedy. The null hypothesis of equal mean is rejected with $p = 4\times 10^{-61}$.
This clearly demonstrate the myopic failure of the greedy policy.

\subsection{MPC+ILQR attack on US Real GNP\label{sec:GNP}}
This real world data~\cite{tiao1994some} models the growth rate of quarterly US real GNP from the first quarter of 1947 to the first quarter of 1991. 
We use the GNP data to evaluate MPC+iLQR and Greedy, which attack the ``last day.''
The environment's nonlinear threshold model dynamics is:
%\begin{small}
\begin{align*}
&x_{t+1} = \\
&
\begin{cases}
-0.015-1.076 x_t + w_{1,t} & (x_t,x_{t-1}) \in X_1  \\
-0.006 + 0.630 x_t - 0.756 x_{t-1} + w_{2,t} \! \! \! \! \!&  (x_t,x_{t-1}) \in X_2\\
0.006 + 0.438 x_t + w_{3,t} & (x_t,x_{t-1}) \in X_3\\
0.004 + 0.443 x_t + w_{4,t} & (x_t,x_{t-1}) \in X_4,
\end{cases}
\end{align*}
%\end{small}  
where $X_1 = \{(x_t,x_{t-1})\mid x_t\le x_{t-1} \le 0\}, X_2 = \{(x_t,x_{t-1})\mid x_t > x_{t-1}, x_{t-1} \le 0\}, X_3 = \{(x_t,x_{t-1})\mid x_t \le x_{t-1}, x_{t-1} > 0\}, X_4 = \{(x_t,x_{t-1})\mid x_t > x_{t-1} > 0\}$, $w_{1,t}\sim N(0,0.0062^2), w_{2,t}\sim N(0,0.0132^2), w_{3,t}\sim N(0,0.0094^2), w_{4,t}\sim N(0,0.0082^2)$. 
We let $T = 10$, $x_0 = 0.0065$ (according to~\cite{tiao1994some}), $x_{-1} = 0$. 
The forecaster's model is $x_{t+1} = 0.0041 + 0.33 x_t + 0.13 x_{t-1} $ (according to~\cite{tiao1994some}). 
%Equivalently, $C = 
%\begin{pmatrix}
%1&0&0\\
%0.0041&0.33&0.13\\
%0&1&0
%\end{pmatrix}$ in matrix notation. 
The attacker can change state value by adding perturbation. % which is directly changing the growth rate of GNP. 
The attack target is to drive forecaster's predictions $y_{T\mid t}, t= 1, \cdots, T-1$ to be close to $0.01$. We let $\tilde \lambda = 0.001$. MPC+iLQR and Greedy are used to solve this problem. The time step of MPC is set to be $l = 5$. Inside the MPC loop, the stopping condition of iLQR is $tol = 10^{-4}$. %$L(\cdot)$ is defined in~\eqref{eq:L}. 
The maximum iteration of iLQR is set to be $1000$. For Greedy, we use the default setting for the \emph{lsqnonlin} solver 
in Matlab~\cite{coleman1999optimization}
except that we do provide the gradients

\begin{figure}[th]
\centering
\includegraphics[width=0.47\textwidth]{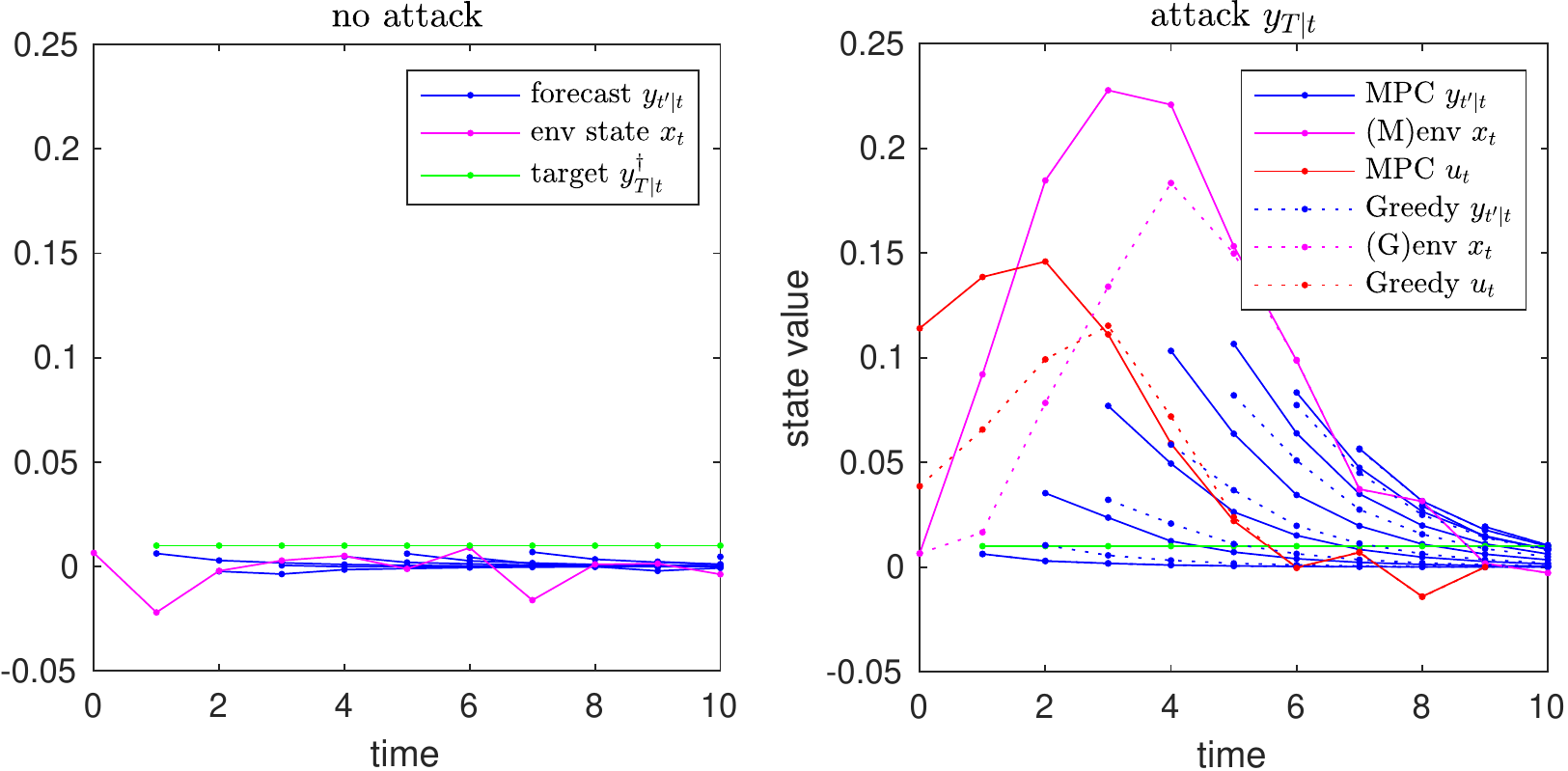}
\caption{MPC+iLQR and Greedy attack on GNP data. MPC+iLQR: $u_4 = 0.0590, y_{T\mid5} = 0.0084$; Greedy: $u_4 = 0.0719, y_{T\mid5} = 0.0079$. $ x_{T \mid 5}^\dagger = 0.01$.\label{fig:real3}}
\end{figure}

We again run 50 trials, the last one is shown in Figure~\ref{fig:real3}.
The mean actual realized cost of no attack, MPC+iLQR attack and Greedy attack are $(6.87,3.03,3.23)\times 10^{-4}$ respectively. The standard errors are $(1.40,0.18,0.19)\times 10 ^{-5}$ respectively. The null hypothesis of equal mean is rejected with $p = 6\times 10^{-65}$ by a paired $t$-test.
As an interesting observation, in the beginning MPC+iLQR adopts a larger attack than Greedy; at time $t = 4$, MPC+iLQR adopts a smaller attack than Greedy, but drives $y_{T\mid 5}$ closer to $0.01$. This shows the advantage of looking into future. Since Greedy only focus on current time, it ignores how the attack will affect the future.
\subsection{Using System Identification in Black-box Attack}
We now show \textbf{system identification can perform black-box attack effectively}. We compare the system identification attack to an oracle, who has full information of both the environmental dynamics and forecast model. The oracle use MPC+ILQR to attack the forecast.

We use the same dynamic in~\cite{fan2008nonlinear} but we change the noise to be $w_t \sim N(0,0.1^2)$. The dynamic is $x_{t+1} = 2 x_t/(1+0.8 x_t^2) + w_t$. We let $x_0 = 3$, $T = 50$. We simulate this dynamical system to $t = 50$ and get a sample sequence from this dynamical system. The forecaster's $AR(1)$ model $C$ is estimated from this sequence. We introduce an attacker who can add perturbation $u_t$ to change the state value: $x_{t+1} = 2 x_t/(1+0.8 x_t^2) + u_t + w_t$. 
$Q$ is ``tomorrow''. The attack target is set to be $y_{t+1 \mid t}^\dagger = 2$. 
We let $\tilde \lambda = 0.01$.

Both the attacker and the oracle do nothing but observe the $x_t$ and $y_{t' \mid t}$ at time $t = 0,\ldots,b+p-2$,  and attack the forecast at time $t = b+p-1, \dots, T-2$.
For system identification, we let $b = 15, l = 5, p = 3$.
% that is the attacker use the last $L_1 = 15$ historical data to estimate a model of order $p=3$ for the environment and a model of order $p=3$ for the forecaster.
For the oracle, the time step of MPC is set to be $l = 10$. 
Inside the MPC loop, the stopping condition of ILQR is $tol = 10^{-4}$. %$L(\cdot)$ is defined in~\eqref{eq:L}. 
The maximum iteration of iLQR is set to be $1000$.

\begin{figure}[th]
\centering
\includegraphics[width=0.47\textwidth]{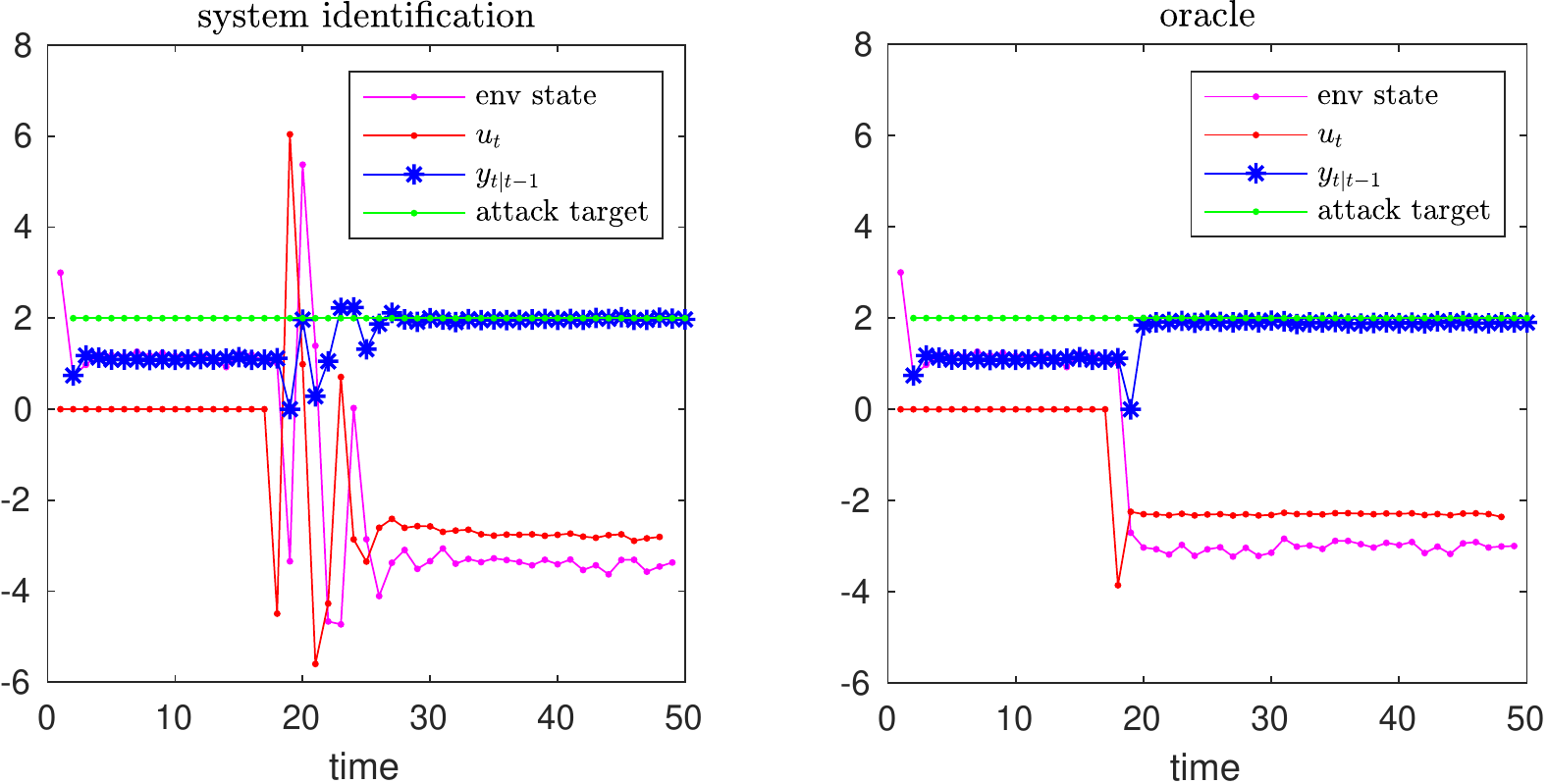}
\caption{System identification and MPC+ILQR oracle attack\label{fig:sysid}}
\end{figure}

We run $100$ trials, the last one is shown in Figure~\ref{fig:sysid}. The mean actual realized cost of system identification attack and oracle MPC+ILQR attack are $4.20, 1.43$ respectively. 
%Even the mean actual realized cost of System identification attack is larger than that of oracle, we still consider the system identification to be an effective attack because system identification requires less knowledge than the oracle.
Even though the cost of System identification attack is larger than that of the oracle, it is evident from Figure~\ref{fig:sysid} that the attacker can quickly force the forecasts (blue) to the attack target (green) after t=25; the chaotic period between t=20 and t=25 is the price to pay for system identification.

%---
%\section{Related Work}

\section{Conclusion}
In this paper we formulated adversarial attacks on autoregressive model as optimal control. 
This sequential attack problem differs significantly from most of the batch attack work in adversarial machine learning.
In the white-box setting, we obtained closed-form LQR solutions when the environment is linear, and good MPC approximations when the environment is nonlinear.
In the black-box setting, we propose a method via system identification then perform MPC.
We demonstrated their effectiveness on synthetic and real data.

%
%The future work includes:
%\begin{itemize} 
%\item Formulate another attack setting: the attacker can only poison forecaster's observation but can not actually change environment state value;
%\item Seek other approximation technique in~\eqref{eq:nmpc} to get a deterministic system when using MPC;
%\item Study attack on autoregressive model with constrained attack space;
%\item Attack RNN using control technique.
%\end{itemize}

%The Observation Attack
%\textbf{Observation attack}: The adversary can only modify the forecaster's observation of the environmental states, but cannot affect the environment state evolution, see Figure~\ref{fig:gm2}.
%We introduce $y_t$ to denote the forecaster's observation of the environment state $x_t$.
%With attack, the forecaster sees $y_t = x_t + u_t$ instead of $y_t=x_t$.
%Now the forecaster's $AR(p)$ model predictions of $\hat x_{t' \mid t}$~\eqref{eq:ARp} is based on $y_t, \ldots, y_{t-p+1}$ instead. 

\section*{Acknowledgment}
We thank Laurent Lessard, Yuzhe Ma, and Xuezhou Zhang for helpful discussions.
This work is supported in part by NSF 1836978, 1545481, 1704117, 1623605, 1561512, the MADLab AF Center of Excellence FA9550-18-1-0166, and the University of Wisconsin.

\bibliography{LQR_AR_Paper,amloc}
\bibliographystyle{aaai}

\appendix
\appendixpage
\addappheadtotoc

\section{A. When $p \neq q$ \label{ap:p!=q}}
For the case when $p \neq q$, we simply inflate the vector dimension. We use $\max(p,q)+1$ dimensional vectors to unify environment model and forecaster's model. Specifically, we redefine the following vectors:
\begin{align}
\bfx_t :=& (1, x_{t}, \ldots, x_{t-\max(p,q)+1})^\top \in \R^{\max(p,q)+1}\\
\hat \bfx_{t' \mid t} :=&
(1,
\hat x_{t' \mid t},
\hat x_{t'-1 \mid t},
\ldots
\hat x_{t'-\max(p,q)+1 \mid t})^\top \\
&\in \R^{\max(p,q)+1}\\
\bar \bfx_{t' \mid t} :=&
(1,
\bar x_{t' \mid t},
\bar x_{t'-1 \mid t},
\ldots
\bar x_{t'-\max(p,q)+1 \mid t})^\top \\
&\in \R^{\max(p,q)+1}
\end{align}
and rewrite the environment dynamics as:
$\bfx_{t+1} = F(\bfx_t, u_t, w_t) := (1, f(x_t, \ldots, x_{t-q+1}, u_t, w_t), x_t, \ldots, x_{t-\max(p,q)+2})^\top.$
Then, redefine $C$ to be the following $(\max(p,q)+1) \times (\max(p,q)+1)$ matrix:
\begin{equation}
C = \begin{bmatrix}
1&0&0&\cdots&0&\cdots&0&0\\
\hat \alpha_0 &\hat \alpha_1 & \hat \alpha_2&\cdots&\hat \alpha_{p} & \cdots & 0 & 0\\
0&1&0&\cdots&0&\cdots&0&0\\
0&0&1&\cdots&0&\cdots&0&0\\
\vdots&\vdots&\vdots&\ddots&\vdots&&\vdots&\vdots\\
0&0&0&\cdots&1&\cdots&0&0\\
\vdots&\vdots&\vdots&&\vdots&\ddots&\vdots&\vdots\\
0&0&0&\cdots&0&\cdots&1&0\\
\end{bmatrix}.
\end{equation}
Thereafter, we can apply LQR for linear $f$ and iLQR for nonlinear $f$. For linear $f$, $P_{t}$ are $(\max(p,q)+1) \times (\max(p,q)+1)$ and $\bfq_t$ are $(\max(p,q)+1) \times 1$.

\section{B. Derivation for Algorithm 1 \label{ap:pfLQR}}
\begin{theorem}\label{thm:LQR}
Under
$
F(\bfx_t,u_t,w_t) = A \bfx_t + B (u_t + w_t)
$
and
$
\beta_{t' \mid t} = 0
$
when $t' \ge t+2$,
the optimal control policy is given by:
\begin{equation}
\phi_t (\bfz) =  -\frac{B^\top \bfq_{t+1} + 2 B^\top P_{t+1} A \bfz}{2(\lambda + B^\top P_{t+1} B)}, \quad t=0 \ldots T-1.
\label{eq:thm1}
\end{equation}
The $(p+1) \times (p+1)$ matrices $P_{t}$ and $(p+1) \times 1$ vectors $\bfq_t$ are computed recursively by a generalization of the Riccati equation:
\begin{align}
P_T	=& 0 \label{eq:thm1P}\\
\bfq_T	=& 0 \label{eq:thm1q}
\end{align}
for $t = T-1, T-2, \cdots,1$ : 
\begin{align}
P_{t} =& C A^\top Q_{t+1 \mid t} C + A^\top (I + \frac{1}{\lambda}P_{t+1} BB^\top)^{-1}P_{t+1} A\\
\bfq_{t} =& -2 C^\top Q_{t+1 \mid t} \bfy_{t+1 \mid t}^\dagger + A^\top \bfq_{t+1} \nonumber \\
&-\frac{1}{\lambda + B^\top P_{t+1} B} A^\top P_{t+1}^\top BB^\top \bfq_{t+1}.
\end{align}
\end{theorem}
This problem can be viewed as an LQR tracking problem with a more complicated cost function and we will show a dynamic programming approach.
\textbf{Proof:}

We will remove the assumption about $\beta_{t' \mid t}$ and give a more general derivation to prove the theorem. It will be straightforward to obtain the exact result in the theorem once we finish the derivation without the assumption about $\beta_{t' \mid t}$.

The dynamic programming has the following steps:
\begin{enumerate}
\item Construct value function sequence $V_t(\cdot)$, for $t = 0,1, \cdots,T$, which can be used to represent the objective function;
\item Solve $\{V_t(\cdot)\}_{t=0}^T$ and the optimal policy $\{\phi_t(\cdot)\}_{t=0}^{T-1}$ by backward recursion.
\end{enumerate}
We will show the detail below:
We first define the \textbf{value function}: for $t = 0,1,\cdots,T$, $V_t(\bfz)$ is:
$$
\min_{u_t,\cdots,u_{T-1}}\E\sum_{\tau=t}^T\sum_{\tau' = \tau+1}^{T}\|C^{\tau'-\tau} \bfx_{\tau} - \bfy_{\tau' \mid \tau}^\dagger\|^2_{Q_{\tau' \mid \tau}}+ \lambda \sum_{\tau=t}^{T-1}u_\tau^2 
$$
where expectation is over $w_t,w_1,\cdots,w_{T-1}$ given $\bfx_t = \bfz$. 
For completeness, $Q_{t' \mid 0}$ is defined to be $0$.
Then $V_0(\bfx_0)$ is the optimal value of objective function. $V_t(\bfz)$ can be found by a backward induction. Note that $V_t(\bfz)$ should always be a quadratic function in $\bfz$. We will show that:
$$
V_t(\bfz) = \bfz^\top P_t \bfz + \bfz^\top \bfq_t + r_t \quad t = 0,1,\cdots,T
$$
where $P_{t}$ is symmetric and positive semi-definite (p.s.d.).

$V_{T}(\bfz) = 0$ can be viewed as a quadratic function with $P_T = 0, \bfq_T = 0, r_T = 0$, $P_{T}$ is symmetric and p.s.d.;

If $V_{t+1}(\bfz) = \bfz^\top P_{t+1} \bfz + \bfz^\top \bfq_{t+1} + r_{t+1}$ and $P_{t+1}$ is p.s.d., we want to show that $V_{t}(\bfz)$ has the same form. By the definition of the value function, we have the following recursive formula for $V_t(\cdot)$:
\begin{align*}
&V_t(\bfz) = \sum_{\tau' = t+1}^T \|C^{\tau'-t} \bfz - \bfy_{\tau' \mid t}^\dagger\|^2_{Q_{\tau' \mid t}}\\
& + \min_v\left(\lambda v^2 + \E_{w_t} V_{t+1} (A \bfz + Bv+ B w_t)\right) \\ 
\end{align*}
Thus,
\begin{align*}
&V_t(\bfz) =  \bfz^\top (\sum_{\tau' = t+1}^T(C^{\tau'-t}) ^\top Q_{\tau' \mid t} C^{\tau'-t} + A^\top P_{t+1} A)\bfz \\
& +  \bfz^\top (-2\sum_{\tau' = t+1}^T(C^{\tau'-t}) ^\top Q_{\tau' \mid t} \bfy_{\tau' \mid t}^\dagger + A^\top \bfq_{t+1}) \\
& + ( \sum_{\tau' = t+1}^T \bfy_{\tau' \mid t}^{\dagger\top} Q_{\tau' \mid t} \bfy_{\tau' \mid t}^\dagger + B^\top P_{t+1} B \sigma^2 + r_{t+1}) \\
& + \min_v ((\lambda + B^\top P_{t+1} B ) v^2 + (B^\top \bfq_{t+1} + 2 B^\top P_{t+1} A \bfz)v ). 
\end{align*}
Since $P_{t+1}$ is p.s.d., $\lambda + B^\top P_{t+1} B \ge \lambda > 0$. Thus, we can solve for the minimum w.r.t. $v$ and obtain the optimal policy function:
\begin{equation}
v^* = \phi_t (\bfz) =  -\frac{B^\top \bfq_{t+1} + 2 B^\top P_{t+1} A \bfz}{2(\lambda + B^\top P_{t+1} B)}.
\end{equation}
Substitute $v$ in $V_t(\bfz)$ with $v^*$ above and we can rewrite $V_t(z)$ as a quadratic function in $z$:
\begin{align*}
&V_t (\bfz) = \bfz^\top (\sum_{\tau' = t+1}^T(C^{\tau'-t}) ^\top Q_{\tau' \mid t} C^{\tau'-t} + A^\top P_{t+1} A \\
& - \frac{1}{\lambda + B^\top P_{t+1} B}A^\top P_{t+1}^\top B B^\top P_{t+1} A)\bfz \\
& + \bfz^\top (-2\sum_{\tau' = t+1}^T(C^{\tau'-t}) ^\top Q_{\tau' \mid t} \bfy_{\tau' \mid t}^\dagger + A^\top \bfq_{t+1} \\
&-\frac{1}{\lambda + B^\top P_{t+1} B} A^\top P_{t+1}^\top BB^\top \bfq_{t+1}) \\
& + ( \sum_{\tau' = t+1}^T \bfy_{\tau' \mid t}^{\dagger\top} Q_{\tau' \mid t} \bfy_{\tau' \mid t}^\dagger + B^\top P_{t+1} B \sigma^2 + r_{t+1}\\
&+ \frac{\bfq_{t+1}^\top B B^\top \bfq_{t+1}}{4(\lambda + B^\top P_{t+1} B)}) \\
\end{align*}
Thus we obtain the following recursive formula for $P_{t}, \bfq_t, r_t$:
\begin{align*}
P_{t} =& \sum_{\tau' = t+1}^T(C^{\tau'-t}) ^\top Q_{\tau' \mid t} ^{\tau'-t} + A^\top P_{t+1} A \\
&- \frac{1}{\lambda + B^\top P_{t+1} B}A^\top P_{t+1}^\top B B^\top P_{t+1} A\\
\bfq_{t} =& -2\sum_{\tau' = t+1}^T(C^{\tau'-t}) ^\top Q_{\tau' \mid t} \bfy_{\tau' \mid t}^\dagger + A^\top \bfq_{t+1} \\
&-\frac{1}{\lambda + B^\top P_{t+1} B} A^\top P_{t+1}^\top BB^\top \bfq_{t+1}\\
r_t =& \sum_{\tau' = t+1}^T \bfy_{\tau' \mid t}^{\dagger\top} Q_{\tau' \mid t} \bfy_{\tau' \mid t}^\dagger + B^\top P_{t+1} B \sigma^2 + r_{t+1} \\
&+ \frac{\bfq_{t+1}^\top B B^\top \bfq_{t+1}}{4(\lambda + B^\top P_{t+1} B)}
\end{align*}
By matrix inversion lemma, $P_t$ can be rewritten as:
$$
\sum_{\tau' = t+1}^T(C^{\tau'-t}) ^\top Q_{\tau' \mid t} ^{\tau'-t} + A^\top (I + \frac{1}{\lambda}P_{t+1} BB^\top)^{-1}P_{t+1} A. 
$$
It's obvious that if $P_{t+1}$ is symmetric and p.s.d., so is $P_{t}$. 

Now, we have completed the backward induction and obtain the optimal control policy $\phi_t(\cdot)$.

\section{C. ILQR \label{ap:pfILQR}}
The problem to be solved is 
\begin{align}\label{eq:nmpc}
\min_{u_\tau,\cdots,u_{L(\tau)}} & \sum_{t=\tau+1}^{L(\tau)+1}\|C \bfx_{t} - \bfy_{t+1 \mid t}^\dagger\|^2_{Q_{t+1 \mid t}} + \sum_{t=\tau}^{L(\tau)}\|u_t\|_R^2 \\
\text{s.t.} & \bfx_\tau \text{ given} \nonumber\\
& \bfx_{t+1} = F(\bfx_t, u_t, 0),\quad t = \tau,\cdots,L(\tau). \label{eq:deter}
\end{align}
Concretely, given $\bfx_\tau$ we initialize the control sequence in some heuristic manner: $\tilde u_\tau, \tilde u_{\tau+1},\cdots,\tilde u_{L(\tau)}$.
If there is not good initialization for the control sequence, just set it to $0$. 
We then simulate the $F$ system in~\eqref{eq:deter} using $\tilde u$ to obtain a state sequence $\tilde \bfx_{\tau}, \tilde \bfx_{\tau+1},\cdots, \tilde \bfx_{L(\tau)+1}$:
\begin{equation}
\tilde \bfx_\tau = \bfx_\tau, \quad \tilde \bfx_{t+1} = F(\tilde \bfx_t,\tilde u_t, 0),\quad t = \tau,\tau+1,\cdots,L(\tau). 
\label{eq:iLQRsimulate}
\end{equation}
While $(\tilde\bfx,\tilde u)$ may not be sensible themselves, they allow us to perform a 
first order Taylor expansion of the nonlinear dynamics $F$ around them:
\begin{align}
\bfx_{t+1} &\approx  F(\tilde\bfx_t, \tilde u_t, 0) + D_\bfx F(\tilde\bfx_t, \tilde u_t, 0) (\bfx_t - \tilde \bfx_t) \nonumber \\
&+ D_u F(\tilde\bfx_t, \tilde u_t,0) (u_t - \tilde u_t) \quad t = \tau, \tau + 1, \cdots, L(\tau),
\end{align}
where $D_\bfx F(\tilde\bfx_t, \tilde u_t, 0)$ and $D_u F(\tilde\bfx_t, \tilde u_t,0)$ are $(p+1)\times(p+1)$ and $(p+1)\times1$ Jacobian matrices:
\begin{equation}
\begin{small}
D_\bfx F = 
\begin{bmatrix}
0&0&0&\cdots&0&0\\
0 &\frac{\partial f}{\partial x_{t}} & \frac{\partial f}{\partial x_{t-1}} &\cdots&\frac{\partial f}{\partial x_{t-p+2}}&\frac{\partial f}{\partial x_{t-p+1}}\\
0&1&0&\cdots&0&0\\
0&0&1&\cdots&0&0\\
\vdots&\vdots&\vdots&\ddots&\vdots&\vdots\\
0&0&0&\cdots&1&0\\
\end{bmatrix}
\nonumber
\end{small}
\end{equation}
\begin{equation}
D_u F(\tilde\bfx_t, \tilde u_t,0)
= (0,\frac{\partial f}{\partial u} ,0, \cdots, 0)^\top.
\nonumber
\end{equation}
Note $F(\tilde\bfx_t, \tilde u_t,0) = \tilde \bfx_{t+1}$ by definition.
Rearranging and introducing new variables 
$\delta\bfx_{t} := \bfx_t - \tilde \bfx_t$,
$\delta u_{t} := u_t - \tilde u_t$,
we have the relation
\begin{equation}
\delta \bfx_{t+1} \approx D_\bfx F(\tilde\bfx_t, \tilde u_t, 0) \delta \bfx_t + D_u F(\tilde\bfx_t, \tilde u_t,0) \delta u_t 
\label{eq:deltaapprox}
\end{equation}
where $t = \tau, \tau + 1, \cdots, L(\tau)$.
We now further approximate~\eqref{eq:nmpc} by substituting the variables and making~\eqref{eq:deltaapprox} \emph{equality} constraints:
\begin{align}
&\min_{\delta u_{\tau:L(\tau)}} \sum_{t=\tau+1}^{L(\tau)+1} \|C (\delta \bfx_t+\tilde \bfx_{t}) - \bfy_{t+1 \mid t}^\dagger\|^2_{Q_{t+1 \mid t}} \nonumber \\
&+ \sum_{t=\tau}^{L(\tau)}\|\delta u_t + \tilde u_t\|_R^2 \label{eq:delLQR}\\
\text{s.t.} & \delta\bfx_\tau = 0 \nonumber\\
& \delta \bfx_{t+1} = D_\bfx F(\tilde\bfx_t, \tilde u_t, 0) \delta\bfx_t + D_u F(\tilde\bfx_t, \tilde u_t,0) \delta u_t. \nonumber
%& \text{  for } t = \tau,\tau+1,\cdots,L(\tau). \nonumber
\end{align} 
The solutions $\delta u$ are then applied as an improvement to $\tilde u_\tau, \tilde u_{\tau+1},\cdots,\tilde u_{L(\tau)}$.
We now have an updated (hopefully better) heuristic control sequence $\tilde u$.
We take $\tilde u$ and iterate the inner loop starting from~\eqref{eq:iLQRsimulate} for further improvement. 

Critically, what enables iLQR is the fact that~\eqref{eq:delLQR} is now an LQR problem (i.e., linear) with a closed-form solution. We show the solution in the following.

\begin{theorem}
In the following, denote $D_uF_s = D_u F(\tilde \bfx_s, \tilde u_s,0), D_\bfx F_s = D_\bfx F(\tilde\bfx_s, \tilde u_s, 0)$. The optimal solution of the LQR problem in each ILQR iteration is given by: for $s = \tau, \cdots, L(\tau)$
\begin{align*}
\delta u_s^* &= -\frac{2 D_u F_s^\top P_{s+1} D_\bfx F_s\delta\bfx_s + D_u F_s^\top \bfq_{s+1} + 2 \lambda \tilde u_s}{2(\lambda + D_u F_s^\top P_{s+1} D_u F_s)}\\
\delta\bfx_{s+1} &= D_\bfx F_s \delta\bfx_s + D_u F_s \delta u_s^*,
\end{align*}
where $\delta\bfx_{\tau} = 0$. The $P_{s}, \bfq_s$ sequences are computed by:
\begin{align*}
	P_{L(\tau)+1} =& C^\top Q_{L(\tau)+2 \mid L(\tau)+1} C\\
	\bfq_{L(\tau)+1} =& 2C^\top Q_{L(\tau)+2 \mid L(\tau)+1} (C \tilde\bfx_{L(\tau)+1}- \bfy_{L(\tau)+2 \mid L(\tau)+1}^\dagger)
\end{align*}
for $s = L(\tau), L(\tau)-1, \cdots, \tau$
\begin{align*}
P_s =& C^\top Q_{s+1 \mid s}C \nonumber \\
&+ D_\bfx F_s^\top(I + \frac{1}{\lambda}P_{s+1} D_u F_s D_u F_s^\top)^{-1}P_{s+1}D_\bfx F_s^\top \\
\bfq_s =& 2C^\top Q_{s+1 \mid s}(C \tilde\bfx_s- \bfy_{s+1 \mid s}^\dagger) + D_\bfx F_s^\top \bfq_{s+1} \\
&-\frac{ (D_u F_s^\top P_{s+1} D_\bfx F_s)^\top (D_u F_s^\top \bfq_{s+1} + 2 \lambda \tilde u_s)  }{\lambda + D_u F_s^\top P_{s+1} D_u F_s}.
\end{align*}

\end{theorem}
\textbf{Proof:}

The proof is similar to the proof of Theorem 1 but there is not noise involved. We will again derive without the assumption about $\beta_{t' \mid t}$. 

We first define the value function: for $s = \tau, \tau+1, \cdots, L(\tau)+1$, $V_s(\bfz)$ is:
\begin{align*}
&\min_{\delta u_s, \cdots, \delta u_{L(\tau)}}\sum_{t=s}^{L(\tau)+1}\sum_{t'=t+1}^T\|C^{t'-t}(\delta \bfx_t+\tilde\bfx_t)-\bfy_{t' \mid t}^\dagger\|^2_{Q_{t' \mid t}} \\
&+ \lambda \sum_{t = s}^{L(\tau)}(\delta u_t + u_t)^2
\end{align*}
subject to $\delta \bfx_{s} = \bfz$. For completeness, $Q_{t' \mid \tau}$ is defined to be $0$. Then $V_\tau(0)$ is optimal value for the objective function. $V_s(\bfz)$ can be found by a backward induction. Note that $V_s(\bfz)$ should always be a quadratic function in $\bfz$. We will show that:
\begin{equation}
V_s(\bfz) = \bfz ^\top P_s \bfz + \bfz^\top \bfq_s + r_s \quad s = \tau, \tau+1, \cdots, L(\tau) + 1
\end{equation}
where $P_s$ is symmetric and positive semi-definite.

First, 
$V_{L(\tau)+1}(\bfz) = \sum_{t' = L(\tau)+2}^T\|C^{t'-(L(\tau)+1)}(\bfz+\tilde\bfx_{L(\tau)+1})-\bfy_{t' \mid L(\tau)+1}^\dagger\|^2_{Q_{t' \mid L(\tau)+1}}$
is a quadratic function with 
\begin{align*}
P_{L(\tau)+1} =& \sum_{t' = L(\tau)+2}^T(C^{t'-(L(\tau)+1)})^\top Q_{t'\mid L(\tau)+1} C^{t'-(L(\tau)+1)}\\
\bfq_{L(\tau)+1} =& 2\sum_{t' = L(\tau)+2}^T(C^{t'-(L(\tau)+1)})^\top Q_{t' \mid L(\tau)+1}\\
& \cdot (C^{t'-(L(\tau)+1)}\tilde\bfx_{L(\tau)+1}-\bfy_{t' \mid L(\tau)+1}^\dagger) \\
r_{L(\tau)+1} =& \sum_{t' = L(\tau)+2}^T(C^{t'-(L(\tau)+1)}\tilde\bfx_{L(\tau)+1}-\bfy_{t' \mid L(\tau)+1}^\dagger)^\top \\
& \cdot Q_{t' \mid L(\tau)+1}(C^{t'-(L(\tau)+1)}\tilde\bfx_{L(\tau)+1}-\bfy_{t' \mid L(\tau)+1}^\dagger).
\end{align*}
Assume $V_{s+1}(\bfz) = \bfz^\top P_{s+1} \bfz + \bfz^\top \bfq_{s+1} + r_{s+1}$, we can write down $V_s(\bfz)$,
\begin{align*}
V_s(\bfz) =& \sum_{t'=s+1}^T\|C^{t'-s}(\bfz +\tilde\bfx_s)-\bfy_{t' \mid s}^\dagger\|^2_{Q_{t' \mid s}}\\
&+ \min_v(\lambda(v+\tilde u_s)^2 + V_{s+1}(D_\bfx F_s \bfz + D_u F_s v))\\
=& \sum_{t'=s+1}^T\|C^{t'-s}(\bfz +\tilde\bfx_s)-\bfy_{t' \mid s}^\dagger\|^2_{Q_{t' \mid s}} \\
&+ (D_\bfx F_s \bfz)^\top P_{s+1} (D_\bfx F_s \bfz) + (D_\bfx F_s \bfz)^\top \bfq_{s+1} \\
& + \lambda \tilde u_s^2 + r_{s+1} + \min_v((\lambda + D_u F_s ^\top P_{s+1} D_u F_s )v^2 \\
&+ (2 D_u F_s ^\top P_{s+1} D_\bfx F_s \bfz + D_u F_s ^\top \bfq_{s+1} + 2 \lambda \tilde u_s)v).
\end{align*}
The optimal control signal is:
\begin{equation}
\delta u_s^* = v^* = -\frac{2 D_u F_s ^\top P_{s+1} D_\bfx F_s \bfz + D_u F_s ^\top \bfq_{s+1} + 2 \lambda \tilde u_s}{2(\lambda + D_u F_s ^\top P_{s+1} D_u F_s)}.
\end{equation}
Substitute $v$ in $V_s(\bfz)$ with the equation above, we get:
\begin{align*}
P_s = & \sum_{t'=s+1}^T(C^{t'-s})^\top Q_{t' \mid s}(C^{t'-s}) + D_\bfx F_s ^\top P_{s+1} D_\bfx F_s \\
& - \frac{( D_u F_s ^\top P_{s+1} D_\bfx F_s)^\top D_u F_s^\top P_{s+1} D_\bfx F_s}{\lambda + D_u F_s^\top P_{s+1} D_u F_s}\\
\bfq_s =& 2\sum_{t'=s+1}^T(C^{t'-s})^\top Q_{t' \mid s}(C^{t'-s} \tilde\bfx_s-\bfy_{t' \mid s}^\dagger) + D_\bfx F_s^\top \bfq_{s+1} \\
&-\frac{ (D_u F_s^\top P_{s+1} D_\bfx F_s)^\top (D_u F_s^\top \bfq_{s+1} + 2 \lambda \tilde u_s)  }{\lambda + D_u F_s^\top P_{s+1} D_u F_s}\\
r_s =& \sum_{t'=s+1}^T(C^{t'-s}\tilde\bfx_s-\bfy_{t' \mid s}^\dagger)^\top Q_{t' \mid s}(C^{t'-s}\tilde\bfx_s-\bfy_{t' \mid s}^\dagger) + \lambda \tilde u_s^2 \\
&+ r_{s+1} -\frac{ (D_u F_s^\top \bfq_{s+1} + 2 \lambda \tilde u_s)^\top D_u F_s^\top \bfq_{s+1} + 2 \lambda \tilde u_s}{4(\lambda + D_u F_s^\top P_{s+1} D_u F_s)}.
\end{align*}
By matrix inversion lemma, we can rewrite $P_s$ as:
\begin{align*}
&\sum_{t'=s+1}^T(C^{t'-s})^\top Q_{t' \mid s}(C^{t'-s}) \\
&+ D_\bfx F_s^\top(I + \frac{1}{\lambda}P_{s+1} D_u F_s D_u F_s^\top)^{-1}P_{s+1}D_\bfx F_s^\top
\end{align*}
It's obvious that if $P_{s+1}$ is symmetric and p.s.d., so is $P_{s}$. 

\section{D. Greedy Policy When $F$ is Linear}
The greedy control policies 
at time $t = 0,1,\cdots,T-1$ are
\begin{equation}
\phi^G_t(\bfx_t) = -\frac{(C B)^\top Q_{t+2 \mid t+1} (C A\bfx_t- {\bfy}_{t+2 \mid t+1}^\dagger )}{\lambda+(C B)^\top Q_{t+2 \mid t+1} (C B)}.
\label{eq:greedy}
\end{equation}
When $t = T-1$, $Q_{T \mid T-1}$ is $0$ and $\phi^G_{T-1}(\bfx_{T-1})=0$.
This proof is to solve the minimum of a quadratic function. We will again given a more general derivation. 

We start with
\begin{align*}
&\E_{w_t}\sum_{t' = t+2}^T\|C^{t'-(t+1)}(A\bfx_t+B u_t + B w_t)-\bfy_{t' \mid t+1}^\dagger\|^2_{Q_{t' \mid t+1}} \\
&+ \lambda u_t^2 \nonumber\\
=& \sum_{t' = t+2}^T\|C^{t'-(t+1)}(A\bfx_t+B u_t)-\bfy_{t' \mid t+1}^\dagger\|^2_{Q_{t' \mid t+1}} + \lambda u_t^2\\
& + \sum_{t' = t+2}^T \|C^{t'-(t+1)} B\|^2_{Q_{t' \mid t+1}}\sigma^2.
\end{align*}
Since the last summation is constant w.r.t. $u_t$,
\begin{align*}
u^G_t &= \phi^G_t(\bfx_t)\\
=&-\frac{\sum_{t' = t+2}^T(C^{t'-(t+1)}B)^\top Q_{t' \mid t+1} (C^{t'-(t+1)}A\bfx_t-\bfy_{t' \mid t+1}^\dagger)}{\lambda+\sum_{t' = t+2}^T(C^{t'-(t+1)}B)^\top Q_{t' \mid t+1} (C^{t'-(t+1)}B)}
\end{align*}
The last equality is because $\lambda > 0$ and $Q_{t' \mid t}$ is p.s.d. for all $(t,t')$ pairs in the sum.

\end{document}